\documentclass{article}

\usepackage[final]{nips_2016}

\usepackage[utf8]{inputenc} 
\usepackage[T1]{fontenc}    
\usepackage{hyperref}       
\usepackage{url}            
\usepackage{booktabs}       
\usepackage{amsfonts}       
\usepackage{nicefrac}       
\usepackage{microtype}      
\usepackage{nicefrac}
\usepackage{graphicx}

\title{Approximating Wisdom of Crowds using K-RBMs}

\author{
  Abhay Gupta \\
  Microsoft India R\&D Pvt. Ltd. \\
  \texttt{abhgup@microsoft.com} \\
}

\begin{document}

\maketitle

\begin{abstract}
An important way to make large training sets is to gather noisy labels from crowds of non experts. We propose a method to aggregate noisy labels collected from a crowd of workers or annotators. Eliciting labels is important in tasks such as judging web search quality and rating products.  Our method assumes that labels are generated by a probability distribution over items and labels. We formulate the method by drawing parallels between Gaussian Mixture Models (GMMs) and Restricted Boltzmann Machines (RBMs) and show that the problem of vote aggregation can be viewed as one of clustering. We use $K$-RBMs to perform clustering. We finally show some empirical evaluations over real datasets.
\end{abstract}

\section{Introduction}
There has been considerable amount of work on learning when labeling is expensive, such as techniques on transductive inference and active learning. With the emergence of crowdsourcing services, like Amazon Mechanical Turk, labeling costs in many applications have dropped dramatically. Large amounts of labeled data can now be gathered at low price. Due to a lack of domain expertise and misaligned incentives, however, labels provided by crowdsourcing workers are often noisy. To overcome the quality issue, each item is usually simultaneously labeled by several workers, and then we aggregate the multiple labels with some manner, for instance, majority voting.

An advanced approach for label aggregation is suggested by Dawid and Skene[1]. They assume that each worker has a latent confusion matrix for labeling. The off-diagonal elements represent the probabilities that a worker mislabels an arbitrary item from one class to another while the diagonal elements correspond to her accuracy in each class. Worker confusion matrices and true labels are jointly estimated by maximizing the likelihood of observed labels. One may further assume a prior distribution over worker confusion matrices and perform Bayesian inference [2][3][4].

The method of Dawid-Skene (1979) implicitly assumes
that a worker performs equally well across all items in a
common class. In practice, however, it is often the case that
one item is more difficult to label than another. To address
this heterogeneous issue, Zhou et al.(2012)[5] propose a minimax entropy principle for crowdsourcing. It results in that each item is associated with a latent confusion vector besides a latent confusion matrix for each worker. Observed labels are determined jointly by worker confusion matrices and item confusion vectors through an exponential family model. Moreover, it turns out that the probabilistic labeling model can be equivalently derived from a natural assumption of objective measurements of worker ability and item difficulty. Such kinds of objectivity arguments have been widely discussed in the literature of mental test theory [6][7]. All the above approaches are for aggregating multiclass labels and  In many scenarios, the labels are ordinal. Zhou et. al. (2014)[8] proposed a work to aggregate votes using minimax conditional entropy for ordinal labels.

Most of the methods use statistical methods to aggregate the observed labels by transforming them to some probability or entropy measures. But, there has been no work that  operates directly on the observed labels. We present a method to learn the aggregates of the votes using clustering. We first show the formulation that allows us to use clustering as an approximation of the vote aggregation stratagem. We first draw a parallel between the Restricted Boltzmann Machine (RBM) learning and the Expectation Maximization (EM) algorithm of the David-Skene algorithm and then show that Gaussian-Softmax RBMs[9] can be approximated by a Gaussian Mixture Model (GMM), whose specific conditions lead to a direct mapping to the traditional K-means algorithm[10][11].

To then elucidate the clustering paradigm, we perform clustering using the $K$-RBM model as proposed in [14].

\section{Related Work}

\subsection{Restricted Boltzmann Machines}
The restricted Boltzmann machine is a bipartite, undirected graphical model with visible (observed) units and hidden (latent) units. The RBM can be understood as an MRF with latent factors that explains the input visible data using binary latent variables. The RBM consists of visible data $v$ of dimension $L$ that can take real values or binary values, and stochastic binary variables $h$ of dimension $K$. The parameters of the model are the weight matrix $W \in R^{LxK}$ that defines a potential between visible input variables and stochastic binary variables, the biases $c \in R^{L}$ for visible units, and the biases $b \in R^{K}$ for hidden units. When the visible units are real-valued, the model is called the Gaussian RBM, and its joint probability distribution can be defined as follows:

\begin{equation}
	P(v,h) = \frac{1}{Z} exp(-E(v,h))
\end{equation}
\begin{equation}
	E(v,h) = \frac{1}{2\sigma^2}||v-c||^2-\frac{1}{\sigma}v^TWh-b^Th
\end{equation}

where Z is a normalization constant. The conditional distribution of this model can be written as:

\begin{equation}
	P(v_i|h) = N(v_i;\sigma \sum_j W_{ij}h_j+c_i, \sigma^2)
\end{equation}
\begin{equation}
	P(h_j=1|v) = sigm(\frac{1}{\sigma}\sum_i{W_{ij} v_i+b_j)}
\end{equation}

where $sigm(s) = \frac{1}{1+exp(-s)}$ is the sigmoid function, and $N (.; ., .)$ is a Gaussian distribution. Here, the variables in a layer (given the other layers) are conditionally independent, and thus we can perform block Gibbs sampling in parallel. The RBM can be trained using sampling-based approximate maximum-likelihood, e.g., contrastive divergence approximation [12]. After training the RBM, the posterior (Equation 2) of the hidden units (given input data) can be used as feature representations for classification tasks.

\subsection{Gaussian-softmax RBMs}
We define the Gaussian-softmax RBM as the Gaussian RBM with a constraint that at most one hidden unit can be activated at a time given the input, i.e., $\sum_j h_j \leq 1$. The energy
function of the Gaussian-softmax RBM can be written
in a vectorized form as follows:

\begin{equation}
	E(v,h) = \frac{1}{2\sigma^2}||v-c||^2-\frac{1}{\sigma}v^TWh-b^Th
\end{equation}

subject to $\sum_j h_j \leq 1$. For this model, the conditional probabilities of visible or hidden units given the other layer can be computed as:

\begin{equation}
\label{gsrbm}
	P(v|h) = N(v;\sigma Wh+c, \sigma^2I)
\end{equation}
\begin{equation}
	P(h_j=1|v) = \frac{exp(\frac{1}{\sigma}w^T_jv+b_j)}{1+\sum_{j'}exp(\frac{1}{\sigma}w^T_{j'}v+b_{j'})}
\end{equation}

where $w_j$ is the $j$-th column of the $W$ matrix, often denoted as a “basis” vector for the $j$-th hidden unit. In this model, there are $K +1$ possible configurations (i.e., all hidden units are 0, or only one hidden unit $h_j$ is 1 for some $j$).

\subsection{K-Means}
The k-means clustering is an unsupervised algorithm that assigns clusters to data points. The algorithm can be written as

\begin{itemize}
\item Randomly choose k cluster centers, $\mu^{(0)} = \mu_1^{(0)}, \mu_2^{(0)}, \cdots, \mu_k^{(0)}$.
\item Assign an incoming data point $x_j$ to the nearest cluster center $C^{(t)}(j) = \min_i ||\mu_i - x_j||^2$
\item $\mu_i$ becomes centroid of the cluster. $\mu_i^{t+1} = \min_{\mu} \sum_{j:C(j)=i}{||\mu-x_j||^2}$
\end{itemize}

The procedure is repeated till convergence, that is all points have been assigned the best cluster centers and over many trials $r$, we take the best possible solution as the cluster assignment.

\subsection{Gaussian Mixture Models}
The Gaussian mixture model is a directed graphical model where the likelihood of visible units is expressed as a convex combination of Gaussians. The likelihood of a GMM with $K+1$ Gaussians can be written as follows:

\begin{equation}
\label{gmm}
	P(v) = \sum_{k=0}^{K} \pi_kN(v;\mu_k,\Sigma_k)
\end{equation}

For the rest of the paper, we denote the GMM with shared
spherical covariance as GMM($\mu_k,\sigma^2I$), when $\Sigma_k = \sigma^2I$ for all $k \in \{0, 1, . . . , K\}$. For the GMM with arbitrary positive definite covariance matrices, we will use the shorthand notation GMM($\mu_k, \Sigma_k$).

K-means can be understood as a special case of GMM with spherical covariance by letting $\sigma \rightarrow 0$ [13]. Compared to GMM, the training of K-means is highly efficient; therefore it is plausible to train K-means to provide an initialization of a GMM. Then the GMM is trained with EM algorithm. The EM algorithm learns the parameters to maximize the likelihood of the data as described by Equation \ref{gmm}.

\section{Vote Aggregation as a Clustering Problem}

The following section outlines the proof for vote aggregation as a special case of clustering problems, when trying to model the problem using RBMs.

\subsection{Vote Aggregation using RBMs}

RBMs are learned in a manner so as to minimize the negative log-likelihood of the data. In a vote aggregation setup, the data is the observed labels. Thus, we can see that learning from RBMs are similar to aggregating votes from the Dawid-Skene algorithm which also minimizes the negative log-likelihood of the observed labels. But in the training of RBMs, we often encounter the normalization constant $Z$, which is intractable and this makes it difficult to train an RBM, and we need to approximate $Z$ to learn the ideal parameters for the same. Hence, it becomes difficult to directly apply RBMs to aggregate votes.

\subsection{Equivalence between Gaussian Mixture Models and RBMs with a softmax constraint}

In this section, we show that a Gaussian RBM with softmax
hidden units can be converted into a Gaussian mixture
model, and vice versa. This connection between mixture
models and RBMs with a softmax constraint completes the
chain of links between K-means, GMMs and Gaussian-softmax
RBMs and helps us to visualize vote aggregation as a clustering problem.

As Equation \ref{gsrbm} shows, the conditional probability of visible units given the hidden unit activations for Gaussian-softmax RBM follows a Gaussian distribution. From this perspective, the Gaussian-softmax RBM can be viewed as a mixture of Gaussians whose mean components correspond to possible hidden unit configurations. In this section, we show an explicit equivalence between these two models by formulating the conversion equations between GMM($\mu_k, \sigma^2I$) with $K+1$ Gaussian components and the Gaussian-softmax RBM with $K$ hidden units.

\textbf{Proposition} The mixture of $K + 1$ Gaussians with
shared spherical covariance of $\sigma^2I$ is equivalent to the Gaussian-softmax RBM with K hidden units. We prove the following conversions by showing the following conversions.

\textbf{1. From Gaussian-softmax RBM to GMM($\mu_k,\sigma^2I$):}

We begin by decomposing the chain rule:
\begin{equation}
	P(v,h) = P(v|h)(h),
\end{equation}

where

\begin{equation}
	P(h) = \frac{1}{Z} \int exp(-E(v,h)) dv
\end{equation}

Since there are only a finite number of hidden unit configurations, we can explicitly enumerate the prior probabilities:

\begin{equation}
	P(h_j=1) = \frac{\int exp(-E(h_j=1,v)) dv}{\sum_{j'} \int exp(-E(h_{j'}=1,v)) dv}
\end{equation}

If we define $\hat{\pi_j} = \int exp(-E(v,h_j=1))dv$, then we have $P(h_j=1) = \frac{\hat{\pi_j}}{\sum_{j'} \hat{\pi_{j'}}} \cong \pi_j$. In fact, $\hat{\pi_j}$ can be calculated analytically,

\begin{eqnarray*}
	\hat{\pi_j} &=& \int exp(-E(v,h_j=1)) dv \\
    			&=& exp(\frac{1}{2\sigma^2}||v-c||^2-\frac{1}{\sigma}v^Tw_j-b_j) \\
                &=& (\sqrt{2\pi} \sigma)^L exp(b_j+\frac{1}{2}||w_j||^2 + \frac{1}{\sigma} c^Tw_j)
\end{eqnarray*}

Using this definition, we can show the equivalence as,
\begin{equation}
	P(v) = \sum_j \pi_jN(v;\sigma w_j+c; \sigma^2I)
\end{equation}

\textbf{2. From GMM($\mu_k, \sigma^2I$) to Gaussian-softmax RBM:}

We will also show this by construction. Suppose we have
the following Gaussian mixture with $K+1$ components and
the shared spherical covariance $\sigma^2I$:
\begin{equation}
	P(v) = \sum_{j=0}^{K} \pi_jN(v;\mu_j,\sigma^2I)
\end{equation}

This GMM can be converted to a Gaussian-softmax RBM with the following transformations:
\begin{equation}
	c = \mu_0
\end{equation}
\begin{equation}
	w_j = \frac{1}{\sigma}(\mu_j-c)
\end{equation}
\begin{equation}
	b = log\frac{\pi_j}{\pi_0}-\frac{1}{2}||w_j||^2-\frac{1}{\sigma}w_j^Tc
\end{equation}

It is easy to see that the conditional distribution $P(v|h_j =1)$ can be formulated as a Gaussian distribution with mean $\mu_j = \sigma w_j+c$, which is identical to that of the Gaussian-softmax RBM.

Further, we can recover the posterior probabilities of the hidden units given visible units as the follows:

\begin{eqnarray*}
	P(h_j=1|v) &=& \frac{\pi_j exp(\frac{-1}{2\sigma^2} ||v-\sigma w_j -c||^2)}{\Pi_{j'=0}^{K} \pi_{j'} exp(\frac{-1}{2\sigma^2} ||v-\sigma w_{j'} -c||^2)} \\
    		   &=& \frac{exp(\frac{1}{\sigma}w^T_jv+b_j)}{1+\sum_{j'}exp(\frac{1}{\sigma}w^T_{j'}v+b_{j'})}
\end{eqnarray*}

Therefore, a Gaussian mixture can be converted to an equivalent Gaussian RBM with a softmax constraint.


\subsection{From GMMs to Clustering Assignments}
GMMs learn a density function over the data, while trying to maximize its likelihood. From maximum likelihood estimation, the equation a GMM tries to learn is $\max \Pi_j P(x_j,y_j)$. But since we do not know $y_j$'s, we maximize the marginal likelihood, which is given by $\max \Pi_j P(x_j) = \max \Pi_j \sum_{i=1}^{k} P(y_j=i, x_j)$ ,where $k$ is the number of clusters.

From the Gaussian Bayes Classifier, $P(y=i|x_j) = \nicefrac{P(x_j|y=i)P(y=i)}{P(x_j)}$, that is,
\begin{equation}
	P(y=i|x_j) \propto \frac{1}{(2\pi)^{\nicefrac{m}{2}}||\Sigma_i||^{\nicefrac{1}{2}}} exp\bigg[-\frac{1}{2}(x_j-\mu_i)^T\Sigma_i^{-1}(x_j-\mu_i)\bigg] P(y=i)
\end{equation}

When $P(x|y=i)$ is spherical, with same $\sigma$ for all classes, $P(x_j|y=i) \propto exp\bigg[\frac{-1}{2\sigma^2}||x_j-\mu_i||^2\bigg]$. If each $x_j$ belongs to one class $C_j$, marginal likelihood is given by:

\begin{equation}
	\Pi_{j=1}^{m}\sum_{i=1}^{k} P(x_j, y=i) \propto \Pi_{j=1}^{m} exp\bigg[\frac{-1}{2\sigma^2}||x_j - \mu_{C_j}||^2\bigg]
\end{equation}

For estimating the parameters, we maximize the log-likelihood with respect to all clusters and this gives,
\begin{equation}
	\max \sum_{j=1}^{m} log(\sum_{i=1}^{k} P(x_j, y=i)) \propto \max \sum_{j=1}^{m} \bigg[\frac{-1}{2\sigma^2}||x_j - \mu_{C_j}||^2\bigg]
\end{equation}

Equivalently, minimizing the negative log-likelihood gives,
\begin{equation}
	\min \sum_{j=1}^{m} -log(\sum_{i=1}^{k} P(x_j, y=i)) \propto \min \sum_{j=1}^{m} \bigg[\frac{1}{2\sigma^2}||x_j - \mu_{C_j}||^2\bigg]
\end{equation}

which is the same as the k-means function. We thus show that the vote aggregation methodolody when applied from an RBM model perspective can be viewed as a clustering problem, one of K-means specifically.

Thus, we can consider vote aggregation learned by maximizing the likelihood of observed labels to be a clustering problem.

\section{Clustering using K-RBMs}
Our framework uses $K$ component RBMs. Each component RBM learns one non-linear subspace. The visible units $v_i$ , $i = 1, \cdots, I$ correspond to an $I$ dimensional visible (input) space and the hidden units $h_j, j = 1, \cdots, J$ correspond to a learned non-linear $J$-dimensional subspace.

\subsection{K-RBM Model}
The $K$-RBM model has $K$ component RBMs. Each of these maps a set of sample points $x_n \in R^I$ to a projection in $R^J$. Each component RBM has a set of symmetric weights (and asymmetric biases) $w_k \in R^{(I+1)x(J+1)}$ that learns a non-linear subspace. Note that these weights include the forward and backward bias terms. The error of reconstruction for a sample $x_n$ given by the $k$th RBM is simply the squared Euclidean distance between the data point $x_n$ and its reconstruction by the $k$th RBM, computed using. We denote this error by $\epsilon_{kn}$. The total reconstruction error $\epsilon_t$ in any iteration $t$ is given by $\sum_{n=1}^{N} \min_k {\epsilon_{kn}}$.

The $K$ RBMs are trained simultaneously. During the RBM training, we associate data points with RBMs based on how well each component RBM is able to reconstruct the data points. A component RBM is trained only on the training data points associated with it. The component RBMS are given random initial weights $w_k, k = 1, \cdots, K$.

\subsection{Methodology}
As in traditional K-means clustering, the algorithm alternates between two steps: (1) Computing association of a data point with a cluster and (2) updating the cluster parameters. In $K$-RBMs, the $n$th data point is associated with the $k$th RBM (cluster) if its reconstruction error from that RBM is lowest compared to other RBMS, i.e. if $\epsilon_{kn} < \epsilon_{k'n} \forall k \neq k', k, k' \in \{1, \cdots, K\}$.

Once all the points are associated with one of the RBMS the weights of the RBMS are learnt in a batch update. In hard clustering the data points are partitioned into the clusters exhaustively (i.e. each data point must be associated with some cluster) and disjointly (i.e. each data point is associated with only one cluster). In contrast with K-means where the update of the cluster center is a closed form solution given the data association with clusters, in $K$-RBMs the weights are learned iteratively.

\section{Experimental Results}
In this section, we report empirical results of our method on
real crowdsourced data. We consider the $L_0$ error metric. Let us denote by $y$ the true rating and $y_b$ the estimate. The error metrics is defined as: (1) $L_0$ = $I(y \neq y_b)$. All the research (code and datasets) is reproducible and is available at: \url{https://github.com/gupta-abhay/deep-voteaggregate}.

\subsection{Data}
\textbf{Web search relevance rating} The web search relevance rating dataset contains 2665 query-URL pairs and 177 workers[5]. Give a query-URL pair, a worker is required to provide a rating to measure how the URL is relevant to the query. The rating scale is 5-level: perfect, excellent, good, fair, or bad. On average, each pair was labeled by 6 different workers, and each worker labeled 90 pairs. More than 10 workers labeled only one pair.

\textbf{Dog Image Labeling} We chose the images of 4 breeds of dogs from the Stanford dogs dataset [8]: Norfolk Terrier (172), Norwich Terrier (185), Irish Wolfhound (218), and Scottish Deerhound (232). The numbers of the images for each breed are in the parentheses. There are 807 images in total. A worker labeled an image at most once, and each image was labeled 10 times.

\subsection{Architectures}
There are four architectures considered for both the datasets. We consider two RBMs, binary-binary RBMs and gaussian-binary RBMs. The architectures are the following:

\subsubsection{Web search relevance rating}
\begin{enumerate}
\item Binary-Binary RBM with 30 visible units and 5 hidden units.
\item Binary-Binary RBM with 18 visible units and 3 hidden units.
\item Gaussian-Binary RBM with 6 visible units and 5 hidden units.
\item Gaussian-Binary RBM with 6 visible units and 3 hidden units.
\end{enumerate}

\subsubsection{Dog Image Labeling}
\begin{enumerate}
\item Binary-Binary RBM with 40 visible units and 4 hidden units.
\item Binary-Binary RBM with 20 visible units and 2 hidden units.
\item Gaussian-Binary RBM with 10 visible units and 4 hidden units.
\item Gaussian-Binary RBM with 10 visible units and 2 hidden units.
\end{enumerate}

\subsection{Results}
We report the results, both $L_0$ and $L_1$ errors of the architectures considered in Tables \ref{web-label} and \ref{dog-label}. The $L_0$  error of the Dawid-Skene model on the web search data is 0.17 and the error on the dog data is 0.21.

\begin{table}[htb!]
\begin{center}
\begin{tabular}{|c|c|c|c|c|}
\hline
Architecture & (30v,5h) & (18v,3h) & (6v,5h) & (6v,3h) \\ \hline
$L_0$ error  &  0.23   &  0.40     &  1.00   &  1.00    \\
\hline
\end{tabular}
\end{center}
\caption{Error metrics for web-search data}
\label{web-label}
\end{table}

\begin{table}[htb!]
\begin{center}
\begin{tabular}{|c|c|c|c|c|}
\hline
Architecture & (40v,4h) & (20v,2h) & (10v,4h) & (10v,2h) \\ \hline
$L_0$ error        &  0.23    &  0.41  &  0.64     &  0.32   \\ \hline
\end{tabular}
\end{center}
\caption{Error metrics for dog-label data}
\label{dog-label}
\end{table}

\subsection{Dicussion and Analysis}
\begin{figure}
\centering
\includegraphics[scale=0.6]{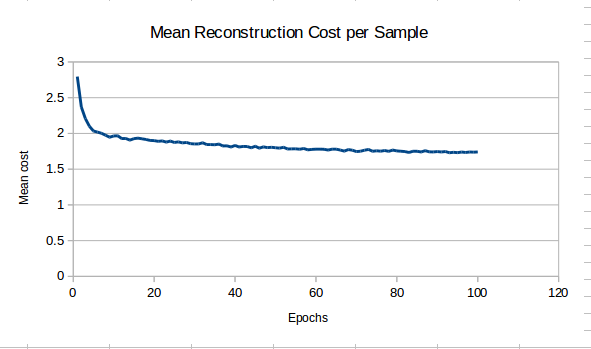}
\caption{Mean reconstruction cost per sameple for the web dataset}
\label{web-sample}
\end{figure}

All the results are done over and average of 20 runs. We see from the results that the results of one-hot encodings are the best among all the proposed architectures, for both the web and dog datasets. This can be because RBMs capture binary data and thus it is able to capture the one-hot encodings in a good manner. Also, we see that in the web data, when we use Gaussian binary RBMs, we get 100\% error. This may be because Gaussian sampling of the data is not ideal for this dataset. On trying CD-k above $k=2$, we get huge reconstruction errors for every data point. However, between CD-1 and CD-2, CD-2 outperforms CD-1. Also, PCD gives huge reconstruction errors for the web dataset, but gave results comparable to CD-1 for the dog dataset. We give a plot for the average reconstruction error per sample as the RBM proceeds for the web dataset in Figure \ref{web-sample}.

\section*{References}
\small

[1] Dawid, A. P. and Skene, A. M. Maximum likeihood estimation of observer error-rates using the EM algorithm. Journal of the Royal Statistical Society, 28(1):20–28, 1979.

[2] Raykar, V. C., Yu, S., Zhao, L. H., Valadez, G. H., Florin, C., Bogoni, L., and Moy, L. Learning from crowds. Journal of Machine Learning Research, 11:1297–1322, 2010.

[3] Liu, Q., Peng, J., and Ihler, A. Variational inference for crowdsourcing. In Advances in Neural Information Processing Systems 25, pp. 701-709, 2012.

[4] Chen, X., Lin, Q., and Zhou, D. Optimistic knowledge gradient policy for optimal budget allocation in crowdsourcing. In Proceedings of the 30th International Conferences on Machine Learning, 2013.

[5] Zhou, D., Platt, J. C., Basu, S., and Mao, Y. Learning from the wisdom of crowds by minimax entropy. In Advances in Neural Information Processing Systems 25, pp. 2204-2212, 2012.

[6] Rasch, G. On general laws and the meaning of measurement in psychology. In Proceedings of the 4th Berkeley Symposium on Mathematical Statistics and Probability, volume 4, pp. 321-333, Berkeley, CA, 1961.

[7] Lord, F. M. and Novick, M. R. Statistical theories of mental test scores. Reading, MA: Addison-Wesley, 1968.

[8] Zhou, D., Liu, Q., Platt, J. and Meek, C. Aggregating ordinal labels from crowds by minimax conditional entropy. In Proceedings of the 31st International Conference on Machine Learning pp. 262-270.

[9] G. E. Hinton and R. Salakhutdinov, “Reducing the dimensionality of data with neural networks,” Science, vol. 313, no. 5786, pp. 504–507, 2006.

[10] Sohn, K., Jung, D.Y., Lee, H. and Hero III, A.O., 2011, November. Efficient learning of sparse, distributed, convolutional feature representations for object recognition. In Computer Vision (ICCV), 2011 IEEE International Conference on (pp. 2643-2650). IEEE.

[11] Andrew Moore Tutorials, \texttt{http://www.autonlab.org/tutorials/}

[12] G. E. Hinton, “Training products of experts by minimizing
contrastive divergence,” Neural Computation, vol. 14, no. 8,
pp. 1771–1800, 2002.

[13] C. Bishop, Pattern recognition and machine learning, vol. 4. Springer New York, 2006

[14] Chandra, S., Kumar, S. and Jawahar, C.V., 2013. Learning multiple non-linear sub-spaces using k-rbms. In Proceedings of the IEEE Conference on Computer Vision and Pattern Recognition (pp. 2778-2785).

\end{document}